%% file: visapp.tex
\documentclass[a4paper,twoside]{article}

\usepackage{epsfig}
\usepackage{subcaption}
\usepackage{calc}
\usepackage{amssymb}
\usepackage{amstext}
\usepackage{amsmath}
\usepackage{amsthm}
\usepackage{multicol}
\usepackage{pslatex}
\usepackage{apalike}
\usepackage{algorithm2e}
\usepackage[bottom]{footmisc}
\usepackage{pdflscape}
\usepackage{SCITEPRESS}     % Please add other packages that you may need BEFORE the SCITEPRESS.sty package.

\usepackage{xcolor}

\newcommand{\PAR}[1]{\vskip2pt \noindent{\bf #1~}}

\begin{document}

\title{Phys-3D: Physics-Constrained Real-Time Crowd Tracking and Counting on Railway Platforms}

\author{\authorname{Bin Zeng\sup{1}, Johannes Künzel\sup{1,2}\orcidAuthor{0000-0002-3561-2758}, Anna Hilsmann\sup{2}\orcidAuthor{0000-0002-2086-0951} and Peter Eisert\sup{1,2}\orcidAuthor{0000-0001-8378-4805}}
\affiliation{\sup{1}Visual Computing Group, Humboldt University Berlin, Unter den Linden 6, 10099 Berlin, Germany}
\affiliation{\sup{2}Fraunhofer Institute for Telecommunications, Heinrich Hertz Institute, HHI, Einsteinufer 37, 10587 Berlin, Germany}
\email{\{johannes.kuenzel, peter.eisert\}@hu-berlin.de, anna.hilsmann@hhi.fraunhofer.de}}


\abstract{Accurate, real-time crowd counting on railway platforms is essential for safety and capacity management. We propose to use a single camera mounted 
in a train, scanning the platform while arriving. While hardware constraints are simple, counting  remains challenging due to dense occlusions, camera motion, and perspective distortions during train arrivals. Most existing tracking-by-detection approaches assume static cameras or ignore physical consistency in motion modeling, leading to unreliable counting under dynamic conditions. We propose a physics-constrained tracking framework that unifies detection, appearance, and 3D motion reasoning in a real-time pipeline. Our approach integrates a transfer-learned YOLOv11m detector with EfficientNet-B0 appearance encoding within DeepSORT, while introducing a physics-constrained Kalman model (Phys-3D) that enforces physically plausible 3D motion dynamics through pinhole geometry. To address counting brittleness under occlusions, we implement a virtual counting band with persistence. On our platform benchmark,MOT-RailwayPlatformCrowdHead Dataset(MOT-RPCH), our method reduces counting error to 2.97\%, demonstrating robust performance despite motion and occlusions. Our results show that incorporating first-principles geometry and motion priors enables reliable crowd counting in safety-critical transportation scenarios, facilitating effective train scheduling and platform safety management.}

\onecolumn \maketitle \normalsize \setcounter{footnote}{0} \vfill

\section{\uppercase{Introduction}}
\label{sec:introduction}
\input{01_Introduction}

\section{\uppercase{Related Work}}
\label{sec:related_work}
\input{02_RelatedWork}

\section{\uppercase{Method}}
\label{sec:methods}
\input{03_Methods}

\section{\uppercase{Experiments}}
\label{sec:experiment}
\input{04_Experiment}

\section{\uppercase{Conclusion}}
\label{sec:conclusion}
\input{05_Conclusion}

\bibliographystyle{apalike}
{\small
\bibliography{visapp}}

\input{Appendix.tex}

\end{document}

%% file: 01_Introduction.tex
\begin{figure}[ht]
    \centering
    \includegraphics[width=0.4\textwidth]{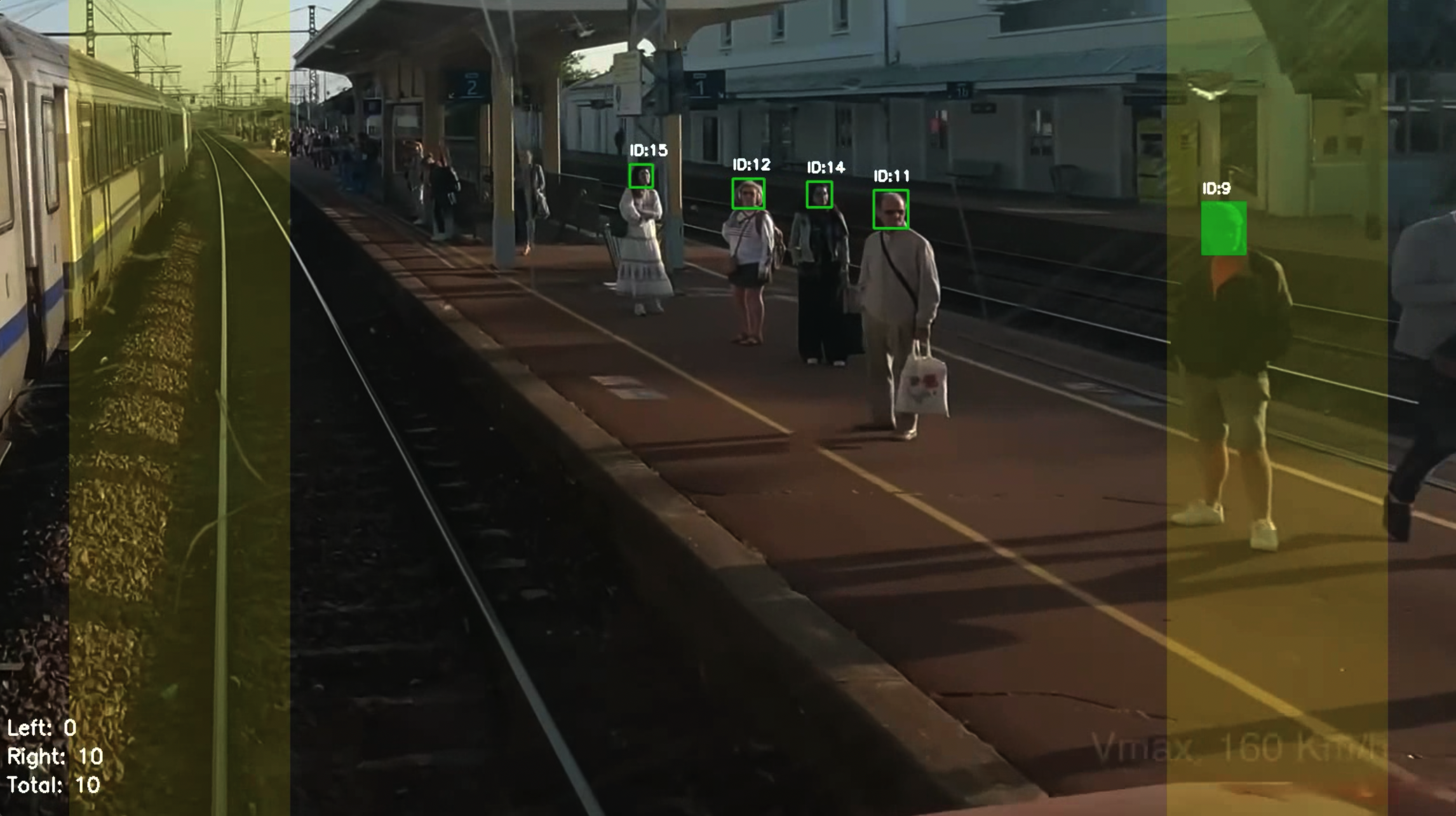}  
    \includegraphics[width=0.4\textwidth]{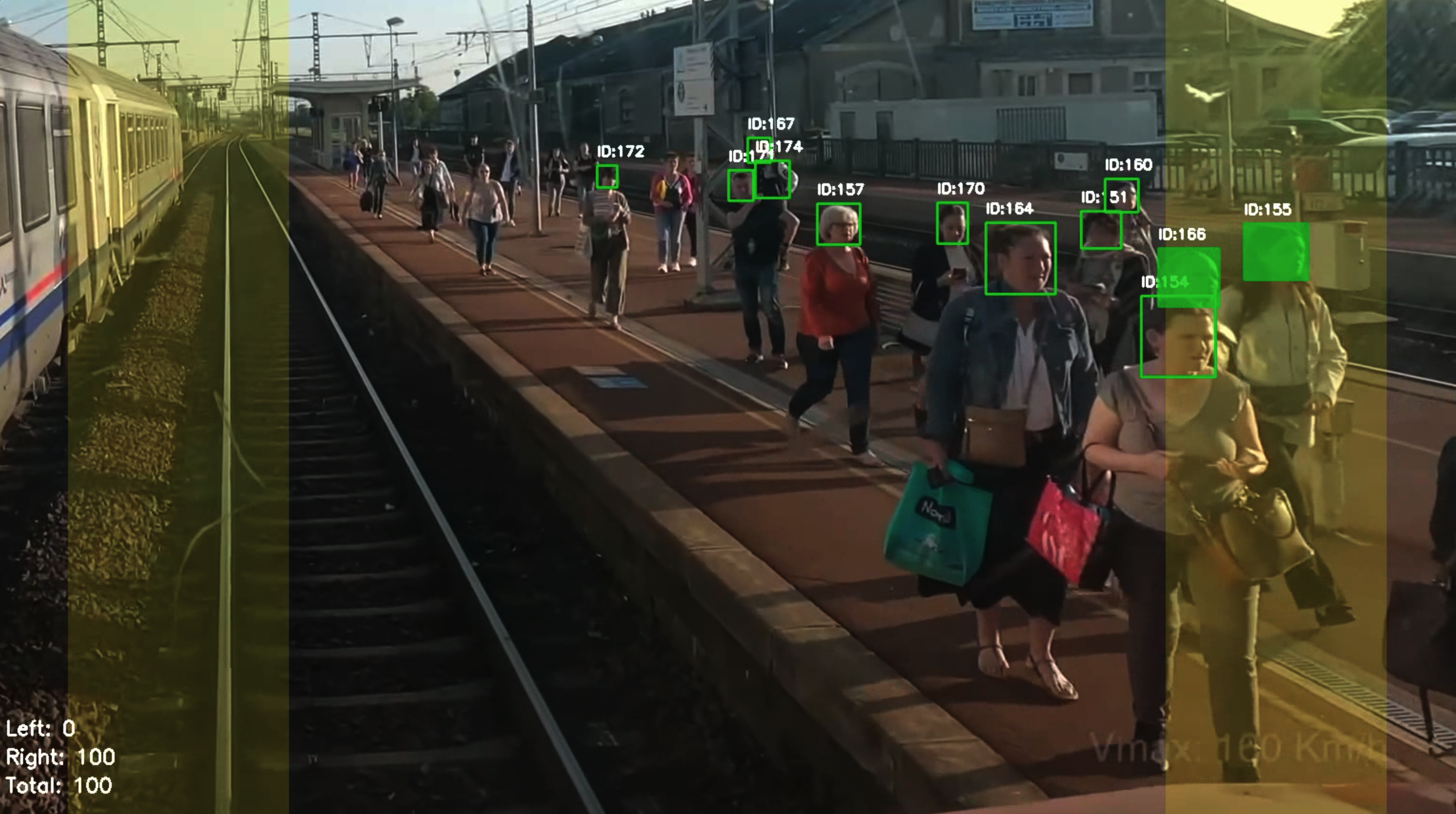}  
    \caption{Two frames from a tracked video sequence. Each bounding box is annotated with a unique track IDs; the bottom-left overlay shows the unique-ID count  from the virtual counting zone (yellow area).}
    \label{fig:intro_preview}
\end{figure}

%For railway operators, obtaining real-time situational awareness of platform crowds during train approach is essential to intelligent dispatch and proactive safety assurance\cite{214,215,216}. A vision system mounted on the train head can provide this "look-ahead" capability, transforming traditional passive platform monitoring into active, forward-looking risk assessment and operational decision-making.
Real-time estimation of passenger density and movement on railway platforms is essential for safe and efficient train operation, as anticipating the number and distribution of waiting passengers enables adaptive dispatching and proactive safety management \cite{214,215,216}. 
Traditional platform surveillance systems rely on static cameras that provide limited coverage and are vulnerable to perspective distortion and occlusion.
Therefore, we aim to enable real-time, onboard perception of platform crowds from a moving train.
Fig.~\ref{fig:intro_preview} shows typical frames from a tracked video sequence showing stable identity tracking and real-time counting results.

%A train-mounted vision system offers a complementary, forward-facing view that supports anticipatory crowd awareness.
However, a moving camera comes with its own specific challenges, as for instance 
%Yet, achieving robust detection, tracking, and counting from a moving camera under dense crowding and strong perspective changes remains a major challenge.
%\subsection{Challenges and Proposed Solutions}
%One major challenge lies in 
detection under severe perspective shifts and mutual occlusions. Distant heads rapidly scale up as the train approaches, and dense crowding often causes extensive overlap among passengers. Full-body detectors (e.g., YOLO variants) suffer degraded performance due to motion blur, heavy overlap, and partial visibility, leading to instability and missed detections~\cite{29_kaur_systematic_2024,30_khan_passenger_2020,01_noauthor_deep_2024,66_zhang_bus_2020,33or34_li_lightweight_2023}. To mitigate these effects, we adopt a head-based detection strategy that leverages the greater visibility and stability of head regions in crowded scenes. We therefore fine-tune a YOLOv11m detector using two-stage transfer learning.
To enable the handling of extreme scale variation and perspective distortion, we extent on the publicly available dataset (CrowdHuman~\cite{shao2018crowdhuman}, Open Sensor Data for Rail 2023~\cite{OpenSensorDataforRail2023}, RailEye3D~\cite{RailEye3D_Dataset}) by meticulously labeling our own domain-specific RailwayPlatformCrowdHead dataset.
%This enables the model to handle extreme scale variation and perspective distortion while maintaining real-time performance.\\
%, consistent with the YOLO family’s wide adoption for efficient visual detection~\cite{02_noauthor_review_nodate,141}.\\

A further challenge arises from ego-motion during the train’s approach. Classical multi-object tracking (MOT) models such as constant-velocity or constant-acceleration Kalman filters~\cite{104,103} often misinterpret camera-induced apparent motion as target motion, leading to physically inconsistent trajectories and frequent identity switches. To address this, we propose a physics-constrained model, termed Phys-3D, which explicitly incorporates the train’s ego-motion into the tracking state prediction. Operating within a perspective-aware coordinate system, Phys-3D decouples true pedestrian movement from camera-induced motion and applies geometric constraints to ensure physically coherent trajectories. This approach enhances temporal stability, reduces identity fragmentation, and preserves computational efficiency through a compact state representation.

Finally, even with stabilized tracks, converting dynamic trajectories into reliable passenger counts presents another layer of difficulty. Short occlusions, detection jitter, and temporary tracking losses can cause duplicate or missing counts when naïve crossing-line methods are applied. To ensure consistent measurement, we define geography-anchored virtual counting regions that correspond to real-world platform zones. By integrating a temporal persistence window, brief interruptions are smoothed out, yielding stable, region-specific counts despite visual disturbances.

Overall, the proposed framework jointly addresses occlusion-robust detection, ego-motion–aware tracking, and stable region-based counting, enabling accurate and reliable passenger analysis in real-world railway environments.
In summary, our main contributions are:
\begin{itemize}
    \item We design a real-time, end-to-end detect–track–analyze pipeline tailored for train approach, enabling forward-looking platform perception onboard the train.
    \item We propose a Phys-3D Kalman filter that incorporates physically grounded ego-motion constraints to address instability under strong perspective and camera motion.
    \item We release a new domain-specific dataset, RailwayPlatformCrowdHead, to facilitate head-based detection and crowd analytics from the train viewpoint.
    \item We demonstrate that combining physics-based motion modeling with deep visual representation yields accurate, stable, and efficient solutions for physically constrained vision tasks in transportation; for tracking-by-detection and DeepSORT foundations see~\cite{132,deepsort,212}.
\end{itemize}

%% file: 02_RelatedWork.tex
Passenger detection and counting have evolved from handcrafted visual models to end-to-end deep learning frameworks. This section reviews related work in three domains: traditional methods, deep learning–based detection and tracking, and railway-specific crowd analysis.\\

Before deep learning dominated the field, crowd and passenger counting relied on handcrafted features and shallow models. Techniques such as Histogram of Oriented Gradients (HOG) and Hough transform were applied to detect passengers in bus or platform images~\cite{30_khan_passenger_2020,73_Huofu}. While computationally efficient, these classical approaches were highly sensitive to illumination, perspective, and occlusion, resulting in poor generalization to crowded or dynamic scenes. Regression-based models that estimated global crowd density from low-level features~\cite{zhang2016single,li2018csrnet} improved robustness but still failed to capture fine-grained individual localization.\\

%\subsection{Deep Learning for Crowd Detection and Counting}
\textbf{Deep Learning for Crowd Detection and Counting.} The advent of convolutional neural networks (CNNs) enabled the transition from feature engineering to feature learning. Early deep regression networks~\cite{zhang2016single,li2018csrnet} predicted pixel-wise density maps, providing accurate crowd estimates in surveillance footage. However, density-based methods lose instance-level information required for multi-object tracking and identity consistency.  
Object detection frameworks such as Faster R-CNN and YOLO series ~\cite{redmon2016you,bochkovskiy2020yolov4,74_yolo11_Jocher_Ultralytics_YOLO_2023} addressed this issue by enabling explicit localization. They have since been adapted to head detection tasks ~\cite{shao2018crowdhuman}, which are more robust in dense or occluded environments. Recent transformer-based detectors further improved long-range context modeling but typically require large datasets and high computational cost, limiting deployment in real-time transportation monitoring.\\

%\subsection{Multi-Object Tracking and Re-Identification}
\textbf{Multi-Object Tracking and Re-Identification}. Multi-object tracking (MOT) extends detection into temporal association. Early online trackers combine Kalman filtering with Hungarian matching, achieving real-time performance~\cite{220}. Later works introduced appearance models and motion prediction enhancements, reducing identity switches in crowded scenarios~\cite{deepsort}. Appearance-based re-identification (Re-ID) has become essential for maintaining consistent identities across frames~\cite{221_OccludedPersonReIdentification_ning2023}. Networks based on EfficientNet and ResNet backbones extract discriminative embeddings that support robust association even under partial occlusions~\cite{222,223_Luo_2020}. Despite these advances, most trackers assume a fixed camera and constant-velocity motion model, which breaks down in scenes with strong perspective variation or moving cameras—typical conditions for railway surveillance.\\

%\subsection{Crowd Counting in Transportation Scenarios}
\textbf{Crowd Counting in Transportation Scenarios}. Compared with generic crowd scenes, railway platforms present unique challenges: frequent partial occlusions, fast camera motion, and pronounced scale variation. While vehicle- or station-level passenger analytics have been studied ~\cite{di_gennaro_hum-card_2024,OpenSensorDataforRail2023}, most approaches rely on full-body detection, making them susceptible to mutual occlusion.  
Recent studies have highlighted the advantages of head-based counting ~\cite{shao2018crowdhuman}, yet publicly available datasets remain limited and lack domain diversity for railway applications. The absence of unified benchmarks and physically grounded motion models further constrains reproducibility and cross-domain generalization.\\

%\subsection{Summary and Research Gap}
In summary, while deep detection and tracking frameworks have achieved impressive accuracy in general crowd scenes, their performance degrades under railway-specific constraints: nonstationary cameras, narrow viewing angles, and dense human occlusion. Traditional constant-velocity Kalman filters fail to model perspective geometry, causing instability in trajectory estimation and erroneous counts. Moreover, few works explicitly address robust and unique counting, separating repeated entries and exits, from a physically constrained motion perspective.  
This motivates our work: an integrated detect-track-count system that combines head-based detection, appearance-aware re-identification, and a novel physics-constrained Kalman model designed to handle perspective distortion and physically plausible motion in real-time railway surveillance.

%% file: 03_Methods.tex
\begin{figure*}
    \centering
    \includegraphics[width=\textwidth]{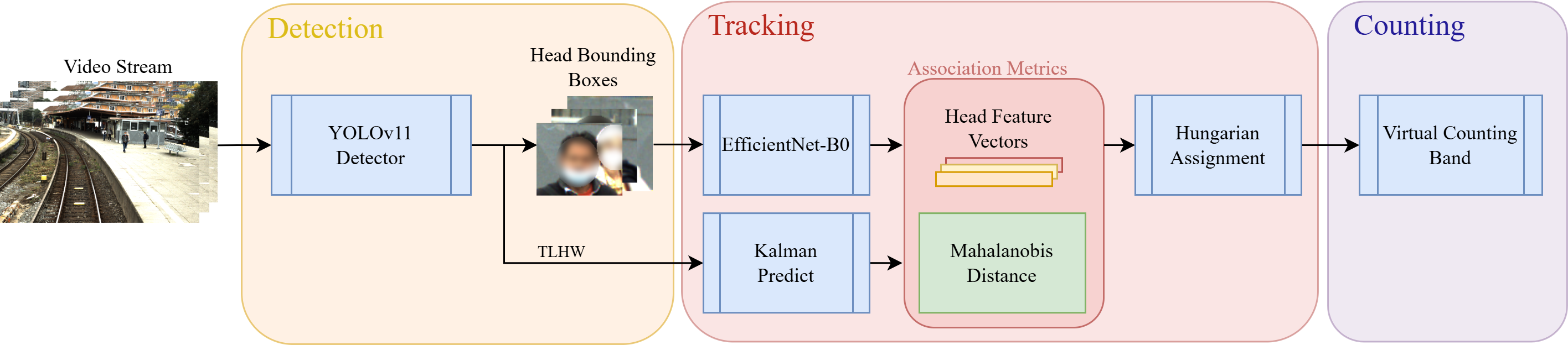}
    \caption{Overview of the proposed physics-constrained detect–track–count pipeline. The Phys-3D model integrates geometric and ego-motion priors to achieve stable identity tracking under camera motion.}
    \label{fig:OurMOTSystemArchitecture}
\end{figure*}

This section presents our approach for accurate crowd counting on railway platforms. 
We propose a physics-constrained detect-track-count pipeline that combines YOLOv11m head detection, EfficientNet-B0-based ReID feature extraction, and a physics-constrained Kalman filter for robust real-time multi-object tracking. 

Our key innovation is the Phys-3D physics-constrained tracker with internal states defined in 3D space instead of the 2D image plane, leveraging camera geometry and scene priors to stabilize identity consistency and achieve superior counting accuracy.

An overview of our proposed system is given in Fig.~\ref{fig:OurMOTSystemArchitecture}.
Incoming video streams are processed in three main stages: detection, tracking, and counting. In the detection stage, a head detector based on YOLOv11m localizes visible pedestrian heads, which provide more reliable cues than full bodies under dense occlusions. 

The tracking stage extends the DeepSORT paradigm by combining visual appearance features from an EfficientNet-B0 encoder with a novel Phys-3D motion model that incorporates 3D geometric constraints and ego-motion priors.
This enables physically consistent trajectory prediction even under strong perspective distortion and train deceleration,
%Specifically, motion prediction is handled by our Phys-3D Kalman model with the state  $[u, v, z, dZ, H]$, representing image-plane coordinates, depth, depth rate, and head height.
as Phys-3D incorporates camera geometry and ego-motion priors to distinguish between true pedestrian motion and apparent motion caused by train movement.

Finally, the counting stage aggregates stable tracks within defined platform regions using virtual counting zones with temporal persistence to handle brief occlusions and detection jitter.

The architecture is modular and runs in real time, allowing onboard crowd perception and analysis during train approach. 
%Fig.~\ref{fig:OurMOTSystemArchitecture} illustrates the overall pipeline and data flow between modules.

\subsection{Head Detection and Encoding}

Accurate detection underpins reliable multi-object tracking in dense crowds.
On railway platforms, full-body detectors fail due to truncation, occlusion, and perspective-induced scale changes.
We therefore focus on head detection, which remains stable and visible even in high-density scenes.

We employ YOLOv11m as the base detector to enhance robustness against occlusion and improve localization of small-scale head targets under strong perspective distortion by pre-training on the CrowdHuman dataset, followed by fine-tuning on our domain-specific RailwayPlatformCrowdHead dataset.
This adapts the model to the unique viewpoint and crowd characteristics of the train-mounted camera, substantially improving localization stability under occlusion and motion blur.
The bounding box of each detected head gets passed to the subsequent tracking module.

%Each detected head is output as a bounding box in TLWH format and passed to the subsequent tracking module along with a cropped image region for appearance feature extraction.\\

We extend the standard DeepSORT \cite{deepsort} with an EfficientNet-B0 to encode detected heads as a 128-dimensional, L2-normalized embedding, enabling re-identification across frames and after temporary occlusions.
EfficientNet-B0 was chosen for its favorable accuracy–efficiency trade-off, which allows real-time deployment on edge hardware and integration into multi-camera systems.
The encoder is trained on a subset of our MOT-RPCH dataset containing 238 identities from seven video sequences. Comprehensive ablations on input resolution and normalization strategies guided the final configuration (128×128 input, aspect-ratio-preserving scaling).
These robust head detections and appearance embeddings form the input to our physics-constrained tracking model, which leverages camera geometry and ego-motion priors to achieve consistent identity tracking under dynamic conditions.

\subsection{Phys-3D: Physics-Constrained 3D Tracking}

Standard tracking-by-detection frameworks such as DeepSORT combine appearance embeddings and 2D motion models effectively by assuming a constant velocity or constant acceleration of the bounding box motion and size in the image plane.
While this works nicely for static video surveillance cameras, tracking might  break down under camera ego-motion, strong perspective distortion and dense occlusions typical for railway platform scenes. In train-mounted cameras, apparent motion in the image plane is dominated by the train’s own motion and deceleration rather than by pedestrian dynamics, causing frequent trajectory drift and identity switches.

To overcome this limitation, we introduce Phys-3D, a physics-constrained 3D Kalman variant, which integrates camera geometry and ego-motion priors into the state prediction to achieve stable and physically consistent trajectory estimation. 

We assume that the dominant motion of the pedestrians in the image plane is due to the ego-motion of the decelerating train along the track.
Instead of modeling bounding box properties in image space, we define the Kalman state in 3D space, which is more more constant. For example, a person's head size typically does not change in 3D while the projection into the image plane varies drastically for a camera approaching the person.

In order to connect the 2D properties of detector and bounding box prediction with the 3D state, we assume a typical pinhole camera
\begin{eqnarray}
x(t) &=& c_x + f_x \cdot \frac{X}{Z(t)} \nonumber \\
y(t) &=& c_y + f_y \cdot \frac{Y}{Z(t)}, 
\label{eq:camera}
\end{eqnarray}
with $f_x,f_y$ being the scaled focal length of the camera, $(x,y)$ being the 2D center of the detected bounding box in image space, and $(X,Y,Z)$ the 3D position of the person's head with respect to the camera coordinate system. For a train approaching the platform on a relatively straight track and mainly in camera viewing direction, $X,Y$ remain almost constant (up to the motion of the people on the platform) while the distance $Z(t)$ changes with the train's position. If the train decelerates with a constant value, velocity and positional updates within a short interval $\Delta t$ can be modeled as
\begin{eqnarray}
Z_{k+1} &=& Z_k + \dot{Z}_k\cdot \Delta t + \frac{1}{2} \cdot \ddot{Z}_k \cdot \Delta t^2 \nonumber \\
\dot{Z}_{k+1} &=& \dot{Z}_k + \ddot{Z}_k \cdot \Delta t\nonumber \\
\ddot{Z}_{k+1} &=& \ddot{Z}_k
\end{eqnarray}

Assuming a constant head height (initialized with $H=0.3m$) in 3D space, distance $Z$ and detected bounding box height $h$ can be related by
\begin{equation}
    Z(t) = f_y \cdot \frac{H}{h(t)}.
\end{equation}
%Due to the approximately cylindrical shape of a head, aspect ratio $a$ in the image plane remains also constant

With these relation between observable 2D quantities and the corresponding 3D values that can be derived through the pinhole camera model eq.~(\ref{eq:camera}), we can setup the Kalman state vector as 
\begin{equation}
    \mathbf{x}_{Phs3D} = [X, Y, H, Z, \dot{Z}, \ddot{Z}]^T.
\end{equation}
This estimated state describes the 3D position of the person $(X,Y,Z)$ with respect to the camera in the train. It assumes a more or less constant position on the platform $(X,Y)$ as well as constant head height $H$, but a changing distance mainly due to the train's ego motion, assuming a constant acceleration $\ddot{Z}$. By modeling the unknowns in 3D space, 2D motion trajectories are more constrained, leading to robust estimates even in case of occlusions.

\subsection{Virtual Counting Band}

Counting individuals in dynamic video streams is challenging due to occlusions, jitter, and identity switches that can cause duplicate or missed counts.
To achieve robust aggregation, we introduce a virtual counting band that maps trajectories to defined spatial regions on the platform.

\begin{figure}[t]
    \centering
    \includegraphics[width=0.49\textwidth]{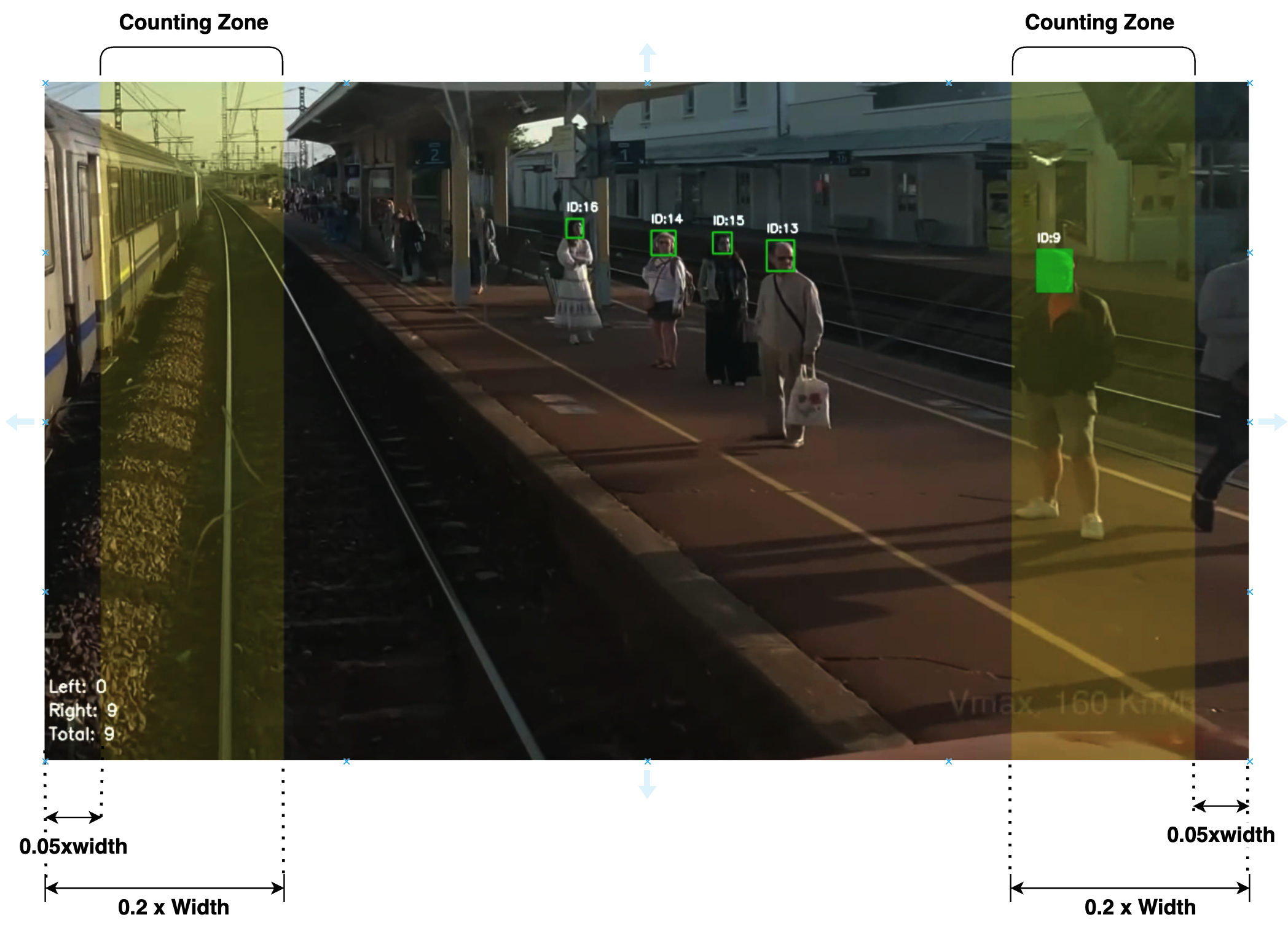}
    \caption{Virtual Counting Band Illustration. The diagram shows the virtual counting zones positioned near the left and right image borders. The band is defined by start and end boundaries as proportions of image width (Start=0.05, End=0.20), creating a buffer region that tolerates brief occlusions and detection jitter. A target is counted only when it remains continuously within the band for a preset number of frames, providing robust counting under challenging conditions.}
    \label{fig:VirtualCountingZone}
\end{figure}

The band is parameterized by its horizontal boundaries ($Start, End$), expressed as proportions of the image width, and by a persistence threshold $N$ defining the required number of consecutive frames a target must remain within the band to be counted.
Each confirmed track maintains its state, including the number of consecutive in-band frames, whether it has already been counted, and its assigned platform side.
Band membership is determined by testing whether the target’s horizontal position lies within the corresponding interval defined by $Start$ and $End$. 
When a track remains within the band for at least N frames, it is counted once, and its ID is stored to prevent duplicate counts. 
At the end of a sequence, tracks partially satisfying the persistence condition can be conditionally included, and counts from both sides are aggregated to yield the final result.
A visual illustration of the image regions can be found in Fig.~\ref{fig:VirtualCountingZone}.

%We assume $H_{\mathrm{head}}=0.3$

%% file: 04_Experiment.tex
This section presents comprehensive experimental evaluation of our Phys-3D physics-constrained tracking system.
We demonstrate the effectiveness of our 3D geometric constraints and physics-based motion modeling for railway platform crowd counting applications.
We start by introducing the used datasets for training and evaluation and giving implementation details to ensure reproducibility.

\subsection{Datasets}
\label{subsec:datasets}

We constructed a comprehensive multi-modal dataset ecosystem for the training and the experiments. 
The following datasets have been used:

\PAR{CrowdHuman} The CrowdHuman dataset~\cite{shao2018crowdhuman} is a large-scale, richly annotated, and highly diverse dataset for detector training.
It contains roughly 470K human instances in 19,370 images.

\PAR{Open Sensor Data for Rail 2023} The OSDaR dataset~\cite{OpenSensorDataforRail2023} comprises multi-sensor sequences.
We sampled 20 images with 180 instances from the RGB videos and manually annotated the headboxes.

\PAR{RailEye3D} The RailEye3D~\cite{RailEye3D_Dataset} was originally intended for training systems targeting passenger safety during the boarding of the train and therefore features video sequences of cameras filming the side of the train.
We manually annotated 320 images with 1,221 instances.

\PAR{RailwayPlatfromCrowd} We also created our own dataset from publicly available YouTube-Videos covering 60 different railway platform scenes.
From these we extracted 1,660 images and manually annotated 5,831 head bounding boxes.

\PAR{MOT-RailwayPlatformCrowdHead} Finally, in 27 video sequences from available YouTube-Videos comprising 24,788 frames, 89,087 bounding boxes, and 885 unique human head identities, we annotated continuous head trajectories.
The number of targets ranges from 3 to 124 and the sequences comprise different illumination, occlusion and crowd density conditions.
The length of the sequences varies between a few seconds and tens of seconds.

\subsection{Implementation Details}

\PAR{Detector} The YOLOv11m detector pre-trained on the CrowdHuman dataset and finetung on a combination of OSDaR (20 images, 180 instances), RailEye3D (320 images, 1,221 instances), and our RailwayPlatformCrowd (1,660 images, 5,831 instances) dataset.
We trained using a SGD optimizer with lr=1e-2, momentum=0.937 and weight decay=5e-4, a cosine annealing scheduler and mixed precision
The training was conducted for 100 epochs for both stages.
Further hyperparameters were kept the same as in the original implementation.
During inference we used a confidence threshold of 0.5, IoU=0.7 and an input size of 640x640.

\PAR{ReID model} We trained the EfficientNet-B0 on 7 sequences with 238 tracked identities from our MOT-RailwayPlatformCrowdHead dataset with an input resolution of 128x128 with aspect-ratio preserving scaling and padding.
The training uses a triplet loss with margin $\alpha=0.5$, AdamW optimizer (lr=$2 \times 10^{-4}$, weight decay=$1 \times 10^{-4}$), OneCycleLR scheduling, mixed precision, and batch size 32.

\PAR{Tracking} We use DeepSORT with three Kalman state-space configurations.
Association uses linear combination of Mahalanobis distance (motion) and cosine distance (appearance) with weight $lambda=0.5$. 
Cascaded matching prioritizes recently observed tracks.
Gating thresholds: CV-8D uses standard DeepSORT thresholds; CA-12D uses stricter
thresholds due to higher dimensionality; Phys-3D uses geometry-aware thresholds based
on depth constraints.

\PAR{Counting} Virtual counting bands with Start=0.05, End=0.20 (relative to image width), persistence threshold N=2 frames. Per-ID de-duplication and end-of-video compensation are applied. Final count uses maximum of left/right counts to avoid non-platform bias.

\PAR{Camera Calibration:} To enable accurate depth estimation and enforce the geometric constraints of the Phys-3D model, all video sequences were camera-calibrated to obtain intrinsic parameters, including focal length and principal point coordinates.
The calibration ensures consistent 3D-to-2D projection accuracy across varying camera resolutions and configurations in the MOT-RPCH dataset.

\PAR{Evaluation} For comprehensive evaluation, we select a representative subset of 20 video sequences from our MOT-RailwayPlatformCrowdHead dataset, spanning multiple resolutions (1536×864 to 2304×1296), frame rates (25, 29.97, 59.94 FPS), and target scales, totaling 18,548 frames, 647 identities, and 73,799 bounding boxes.

\PAR{Metrics} We evaluate using both standard multi-object tracking (MOT) and counting metrics. 
For tracking evaluation, we follow the CLEAR-MOT protocol~\cite{bernardin2008evaluating}, adopting Multiple Object Tracking Accuracy (MOTA), Multiple Object Tracking Precision (MOTP), Identity F1 Score (IDF1), and Identity Switches (IDSW). 
Additionally, we report identity-based metrics, including Identity Precision (IDP), Identity Recall (IDR), and IDF1, as defined in~\cite{ristani2016performance}.
For counting performance assessment, we employ regression-based metrics—Mean Absolute Error (MAE), Root Mean Squared Error (RMSE), Mean Absolute Percentage Error (MAPE), and Mean Error (ME)—to quantify discrepancies between predicted and ground-truth counts across all test videos.

%\textbf{Training Details:} The YOLOv11m detector is pre-trained on CrowdHuman and fine-tuned on 2 k platform images using SGD (lr=0.01, momentum=0.937, weight decay=0.0005) with cosine-annealing scheduling.
%The EfficientNet-B0 encoder for appearance modeling is trained with triplet with margin $\alpha=0.5$, AdamW optimizer (lr=$2 \times 10^{-4}$, weight decay=$1 \times 10^{-4}$), OneCycleLR scheduling, mixed precision, and batch size 32.\\

%%%%%%%%

\subsection{Effectiveness of the Two-Stage Detector Training}

We first pre-train on the larger general crowd dataset (CrowdHuman) and then fine-tune the network for the railway platform domain using the combination of our smaller, but domain-specific datasets.
The performance of the head detector can be seen in Tab.~\ref{tab:yolo_training_results}.
% Reviewer 1 meinte das müsste Tab. 5 sein. Habe desegen "head" ergänzt.
The results of the pretraining are shown in the first row of Tab.~\ref{tab:yolo_training_results} and lead to a precision of  mAP50=79.4\% and mAP50–95=54.7, respectively.
After finetuning, we achieve mAP50=98.0\% (+18.6 points) and mAP50–95=81.6\% (+26.9 points), highlighting the importance of the domain-specific training data.

\begin{table*}[htb]
\caption{Performance of the YOLOv11m network. Upper row: after pre-training on the CrowdHuman Dataset. Lower row: fine-tuning results on the Custom Head Dataset.}
\centering
%\resizebox{0.48\textwidth}{!}{%
%\resizebox{0.95\textwidth}{!}{%
%\setlength{\tabcolsep}{12pt}
\small
\renewcommand{\arraystretch}{1.3}
\begin{tabular}{|c|c|c|c|c|c|c|}
\hline
\textbf{Dataset} & \textbf{Images} & \textbf{Instances} & \textbf{Box(Precision)} & \textbf{Recall} & \textbf{mAP50} & \textbf{mAP50–95} \\
\hline
CrowdHuman & 4370 & 98568 & 0.862 & 0.715 & 0.794 & 0.547 \\
\hline
Custom Human & 600 & 4779 & 0.965 & 0.946 & 0.980 & 0.816 \\
\hline
\end{tabular}%
%}
\label{tab:yolo_training_results}
\end{table*}

\subsection{EfficientNet-B0 ReID Model}

\begin{figure}[htb]
    \centering
    \includegraphics[width=0.48\textwidth]{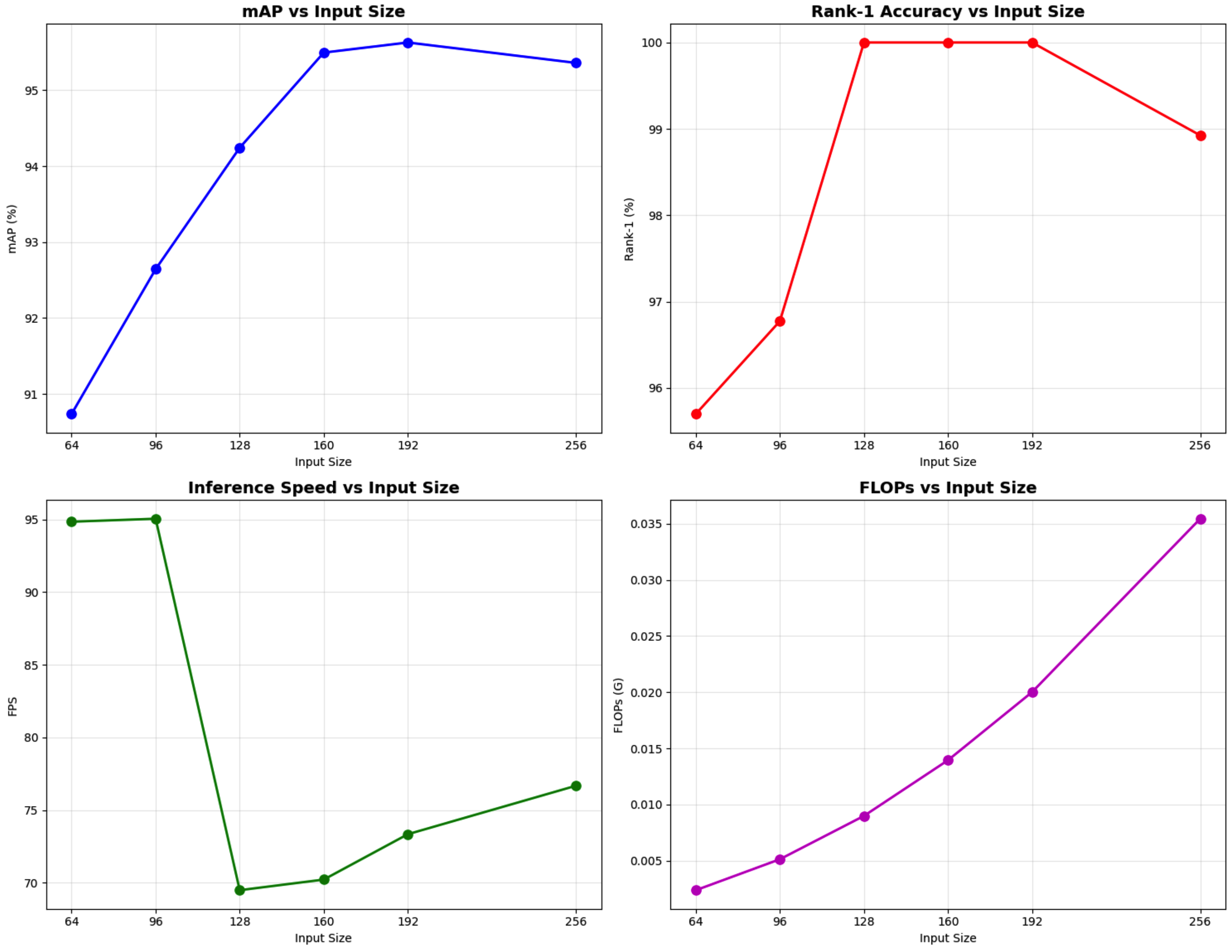}
    \caption{ReID Input Size Ablation Experiment Results}
    \label{fig:ReIDInputsizeAblationexperiment}
\end{figure}

To determine the optimal input resolution for our ReID model, we conduct a comprehensive ablation study comparing multiple input sizes under identical training settings.
For the seven training sequences, we evaluate the speed-accuracy trade-off across multiple resolutions.
We evaluate six different input resolutions from 64×64 to 256×256 pixels in Fig.~\ref{fig:ReIDInputsizeAblationexperiment}.

The results in Fig.~\ref{fig:ReIDInputsizeAblationexperiment} and Tab.~\ref{tab:ReIDablationexperiment} demonstrate clear performance improvements with increasing resolution, but with diminishing returns and increased computational cost.
We observed an apparent FPS inversion at 128×128, where larger inputs (e.g., 160×160) ran marginally faster, attributable to measurement artifacts: batch-size=24 placed 128×128 at a compute–communication tipping point, cuDNN’s Winograd path introduced extra transform overhead, and silent batch-size reductions at larger inputs increased effective GPU boost frequencies.
This inversion does not change our conclusion that 128×128 lies on the accuracy–efficiency Pareto frontier (Rank-1=100\% and mAP=94.24\% at only 9 MFLOPs, leading to 69.5 FPS inference on a NVIDIA T4 GPU), and we therefore deploy 128×128 to ensure reproducibility and avoid confounders.

\begin{table*}[htb]
    \centering
    \caption{ReID Ablation Experiment Results.}
%    \resizebox{0.48\textwidth}{!}{%
    %\resizebox{0.95\textwidth}{!}{%
    \small
    \renewcommand{\arraystretch}{1.3}
    \begin{tabular}{|c|c|c|c|c|c|c|}
    \hline
        \textbf{Input Size (pix)} & \textbf{mAP (\%)} & \textbf{Rank-1 (\%)} & \textbf{FPS} & \textbf{FLOPs (G)} & \textbf{Parameter quantity} & \textbf{Marginal benefits} \\ \hline
        64x64 & 90.74 & 95.7 & 94.8 & 0.002 & 206240 & 0 \\
        96x96 & 92.64 & 96.77 & 95 & 0.005 & 206240 & +1.90 mAP \\
        128x128 & 94.24 & 100 & 69.5 & 0.009 & 206240 & +1.60 mAP \\
        160x160 & 95.49 & 100 & 70.2 & 0.014 & 206240 & +1.25 mAP \\
        192x192 & 95.63 & 100 & 73.3 & 0.02 & 206240 & +0.14 mAP \\
        256x256 & 95.36 & 98.92 & 76.7 & 0.035 & 206240 & –0.27 mAP \\ \hline
    \end{tabular}%
    %}
    \label{tab:ReIDablationexperiment}
\end{table*}

%\textbf{Key Findings:}
%\begin{itemize}
%    \item \textbf{Performance Improvement:} Increasing resolution from 64×64 to 128×128 improves mAP by 3.50\% and Rank-1 by 4.30\%, representing significant accuracy gains.
%    \item \textbf{Diminishing Returns:} Beyond 128×128, performance improvements become marginal (+1.25 mAP for 160×160, +0.14 mAP for 192×192).
%    \item \textbf{Computational Efficiency:} 128×128 provides the best balance, achieving Rank-1=100\% with only 0.009 GFLOPs and 69.5 FPS inference on NVIDIA T4 GPU.
%    \item \textbf{Performance Degradation:} 256×256 shows slight performance degradation (mAP drops by 0.27\%) while requiring significantly more computation.
%\end{itemize}

%Based on this comprehensive analysis, we select 128×128 as our final ReID input resolution, achieving optimal performance with Rank-1=100\%, mAP=94.24\%, and efficient real-time inference capability.

\subsection{Comparative Analysis of Three State-Space Motion Models}

To assess the role of different tracking approaches in MOT, we conduct a comprehensive comparison of three distinct tracking models: (i) the 8D constant-velocity baseline (CV-8D) with standard DeepSORT parameters, (ii) a 12D constant-acceleration model (CA-12D) with adjusted noise modeling for higher dimensionality, and (iii) our 3D physics-constrained model (Phys-3D) with specialized camera calibration and physics-based constraints.
Both Kalman models 8D and 12D use an observation vector $\mathbf{z}=[x, y, a, h]$, with $x,y$ being the center in pixels and $a, h$ the aspect ratio and height of the bounding box, respectively. 
The evaluation was conducted on our evaluation dataset, MOT-RPCH.
For the 8D model, a standard constant velocity is assumed, leading to the state vector

\begin{equation}
    \mathbf{x}_{8D}=[x,y,a,h, \dot{x}, \dot{y}, \dot{a}, \dot{h}]
\end{equation}

while the 12D model assumes a constant velocity, better capturing deceleration during arrivals:

\begin{equation}
    \mathbf{x}_{12D}=[x,y,a,h, \dot{x}, \dot{y}, \dot{a}, \dot{h}, \ddot{x}, \ddot{y}, \ddot{a}, \ddot{h}]
\end{equation}

We evaluate the three distinct tracking models, each configured solely with the adaptations necessary for its own state vector and dynamics, while holding the detector, ReID, association logic, and counting strategy exactly the same.
Tab.~\ref{tab:3modelsCountingComparisonResults} compares their counting performance under these settings.

\begin{table}[ht]
    \caption{Counting performance comparison of the three models. Our Phys-3D approach clearly outperforms standard constant velocity or acceleration models.}
    \centering
    %\resizebox{0.48\textwidth}{!}{%
    \small
    \begin{tabular}{|c|c|c|c|c|}
    \hline
        \textbf{Model} & \textbf{MAE } & \textbf{RMSE} & \textbf{MAPE(\%)} & \textbf{ME } \\ \hline
        \textbf{CV-8D} & 3.4 & 6.1514 & 14.592 & 1.4  \\ 
        \textbf{CA-12D} & 2.4 & 4.5837 & 6.99 & 0.7  \\ 
        \textbf{Phys-3D} & \textbf{0.9} & \textbf{1.3601} & \textbf{2.97} & \textbf{-0.2} \\ \hline
    \end{tabular}%
    %}
    \label{tab:3modelsCountingComparisonResults}
\end{table}

The results indicate that counting accuracy in our platform scene is driven by physically plausible depth/scale evolution and robust association under occlusion.
The CV-8D baseline, which assumes constant image-plane velocity, fails to capture the pronounced arrival deceleration and perspective-driven magnification, leading to larger errors and a positive ME (double counting from ID fragmentation). 
CA-12D partially alleviates this by modeling acceleration, but its higher-dimensional kinematics without geometry/physics constraints remains noise-sensitive, yielding residual overcount ($ \mathrm{ME} > 0 $).
Phys-3D integrates a depth-aware state, an adaptive deceleration prior, and geometry-aware gating, which together enforce monotonic, bounded scale/depth evolution and suppress implausible re-associations.
Combined with EfficientNet-B0 ReID, this reduces ID switches and re-entries through the counting band to achieve the lowest MAE/RMSE and MAPE.
The small negative ME for Phys-3D suggests a conservative bias that prevents double counts at the cost of rare misses, which is preferable for deployment. 
Since detector, ReID, association, and counting settings are held fixed, these gains are attributable to the motion model and its physics-informed constraints rather than confounding factors.

\subsection{Tracking and Counting Performance}

We evaluate the counting performance of our proposed system on our evaluation dataset, MOT-RPCH.
As described, the dataset provides a diverse set of weather conditions, number of persons and sequence lengths. 
As summarized in Tab.~\ref{tab:Phys5D_Performance}, Phys-3D achieves a MOTA of 67.19\% and an IDF1 of 76.32\%, with an average of only 24.5 identity switches. 
The model exhibits high precision (89.06\%) and stable identity preservation under challenging railway platform conditions.

In terms of counting accuracy, as shown in the last row of Tab.~\ref{tab:3modelsCountingComparisonResults}, the Phys-3D system attains a Mean Absolute Percentage Error (MAPE) of 2.97\%, with MAE=0.9 and RMSE=1.36, validating the reliability of the physics-constrained 3D tracking approach for bidirectional passenger counting.

\begin{table}[h]
\centering
\caption{Overall multi-object tracking performance of the Phys-3D model.}
%\resizebox{0.48\textwidth}{!}{%
\tiny
\renewcommand{\arraystretch}{1.3}
\begin{tabular}{|l|c|c|c|c|c|c|}
\hline
\textbf{Metric} & \textbf{MOTA} & \textbf{IDF1} & \textbf{IDSW} & \textbf{Precision} & \textbf{Recall} & \textbf{FAF} \\
\hline
Phys-3D & 67.19\% & 76.32\% & 24.5 & 89.06\% & 77.43\% & 0.32 \\
\hline
\end{tabular}%
%}
\label{tab:Phys5D_Performance}
\end{table}

%\subsubsection{Per-Video Analysis}

%Detailed per-video results and visualizations are provided in the Appendix A, illustrating consistent tracking stability and counting accuracy across diverse resolutions, crowd densities, and lighting conditions. 
These results confirm the robustness and generalization capability of the Phys-3D framework in complex railway platform scenarios.
Our Phys-3D model demonstrates strong performance across both tracking and counting tasks on the MOT-RPCH evaluation dataset. 

\subsection{Ablation Studies}

To validate the effectiveness of individual components in our Phys-3D system, we conduct comprehensive ablation studies on the detector, ReID model, and counting method.

\subsubsection{YOLOv11m Detector Ablation Study}
We design our ablation study to examine the fine-tuned detector's sensitivity to hyperparameters.
We vary three key hyperparameters, input size, confidence threshold, and IoU threshold, according to Tab.~\ref{tab:Parameter_comb_yolo} yielding 64 combinations.

\begin{table}[ht]
\centering
\caption{YOLO detection parameter values tested, yielding 64 combinations.}
\small
\begin{tabular}{|c|c|c|c|c|}
\hline
\textbf{Confidence} & 0.3 & 0.5 & 0.7 & 0.9 \\
\textbf{IoU} & 0.3 & 0.5 & 0.7 & 0.9 \\
\textbf{Image Size} & 640 & 736 & 832 & 960 \\
\hline
\end{tabular}
\label{tab:Parameter_comb_yolo}
\end{table}
    
Through comprehensive evaluation across all combinations of parameters, we identify the optimal configuration that achieves the best balance between detection accuracy and inference speed, given in Tab.~\ref{tab:YOLO11m_Optimal_Config}.

\begin{table}[htb]
    \centering
    \caption{Optimal YOLOv11m detector configuration.}
    %\resizebox{0.48\textwidth}{!}{%
    \tiny
    \setlength{\tabcolsep}{2pt} % default is 6pt
    \begin{tabular}{|c|c|c|c|c|c|c|c|}
    \hline
        \textbf{Conf} & \textbf{IoU} & \textbf{Imgsz} & \textbf{mAP50} & \textbf{mAP50-95} & \textbf{Prec.} & \textbf{Recall} & \textbf{Inf. Time(ms)} \\
    \hline
        0.3 & 0.3 & 736 & 0.9224 & 0.6871 & 0.9193 & 0.8707 & 16.86 \\
    \hline
    \end{tabular}%
    %}
    \label{tab:YOLO11m_Optimal_Config}
\end{table}

The best performing parameter set balances precision–recall and feature resolution–compute trade-offs.
%\todo{this is somehow different from number given in implementation details}
A lower confidence threshold preserves more potential positives, an intermediate IoU threshold reduces duplicates without excessive suppression, and 736 px affords sufficient spatial detail for small heads while controlling cost.
The configuration notably improves mAP50–95, indicating robust performance across IoU thresholds and suitability for real-world deployment.
%The optimal configuration achieves mAP50=92.24\% and mAP50-95=68.71\% with precision=91.93\% and recall=87.07\%, while maintaining reasonable inference time of 16.86ms. This configuration demonstrates the effectiveness of moderate confidence and IoU thresholds combined with 736-pixel input resolution for railway platform head detection. 
%The complete ablation study results across all 64 parameter combinations are provided in the Appendix for detailed analysis.

\subsubsection{Ablation Study of the Counting Method}

In addition, we evaluate the proposed Virtual Counting Band and its hyperparameters as shown in Fig.~\ref{fig:VirtualCountingZone}.
The band introduces non-zero width and a persistence criterion to overcome line-crossing brittleness under jitter and occlusion. We focus on the start and end positions.
We varied $Start \in \{ 0.05, 0.1, 0.15, 0.2 0.3\}$ and $End \in \{ 0.15, 0.2 0.25, 0.3, 0.35, 0.4\}$ with $Start < End$ for the virtual counting band configurations and systematically analyze the impact of start and end positions on counting performance. 
Our analysis reveals that the optimal configuration (Start=0.05, End=0.2) achieves exceptional counting accuracy with minimal systematic bias.

\begin{table}[ht]
    \caption{Optimal virtual counting band configuration (Start=0.05, End=0.2).}\centering
    %\resizebox{0.48\textwidth}{!}{%
    \small
    \setlength{\tabcolsep}{3pt} % default is 6pt
    \begin{tabular}{|c|c|c|c|c|c|}
    \hline
        \textbf{Start} & \textbf{End} & \textbf{MAE} & \textbf{RMSE} & \textbf{MAPE(\%)} & \textbf{ME} \\
    \hline
        0.05 & 0.20 & 0.90 & 1.38 & 2.97 & -0.20 \\
    \hline
    \end{tabular}%
    %}
    \label{tab:Optimal_Counting_Band}
\end{table}

%The optimal configuration (Start=0.05, End=0.20) achieves exceptional counting performance with MAE=0.90, RMSE=1.38, MAPE=2.97\%, and ME=-0.20, indicating near-unbiased counting with minimal variance. This configuration demonstrates the effectiveness of a moderate-width counting band positioned near the image borders, providing robust counting under challenging conditions while maintaining high accuracy. 
%The complete ablation study results across all parameter combinations are provided in the Appendix~C for detailed analysis.

Tab.~\ref{tab:ComparisonofAverage} further contrasts average metrics for different counting strategies: line vs. band based crossing detection. The band dramatically outperforms the line based technique across MAE, RMSE, MAPE, and ME, demonstrating superior robustness and accuracy.

\begin{table}[ht]
    \centering
    \caption{Comparison of average counting metrics lines vs.~zones.}
    \small
    \setlength{\tabcolsep}{3pt} % default is 6pt
    %\resizebox{0.45\textwidth}{!}{%
    \begin{tabular}{|l|c|c|c|c|}
    \hline
        \textbf{Method} & \textbf{MAE} & \textbf{RMSE} & \textbf{MAPE(\%)} & \textbf{ME} \\ \hline
        \raggedright Line-crossing & 31.28 & 43.66 & 93.43 & -31.28 \\ \hline
        \raggedright Counting zones & 3.13 & 4.75 & 10.87 & 0.81 \\ \hline
    \end{tabular}%
    %}
    \label{tab:ComparisonofAverage}
\end{table}

%\subsection{Summary}

%Our comprehensive evaluation of the Phys-3D system demonstrates:

%\textbf{Superior Counting Accuracy:} Phys-3D achieves MAPE=2.24\%, representing a significant improvement over baseline methods (CV-8D: 14.59\%, CA-12D: 6.99\%) and demonstrating the effectiveness of physics-constrained 3D tracking.

%\textbf{Robust Tracking Performance:} The system maintains high tracking accuracy (MOTA=67.21\%, IDF1=76.29\%) with low identity switches (24.8 average), indicating effective identity preservation under challenging conditions.

%\textbf{Component Effectiveness:} Ablation studies confirm the importance of YOLOv11m detector optimization, EfficientNet-B0 ReID backbone selection, and virtual counting band design for overall system performance.

%\textbf{Real-time Capability:} The complete system processes at 29.7 FPS on NVIDIA T4 GPU, meeting practical deployment requirements for railway platform monitoring.

These results establish Phys-3D as a state-of-the-art solution for railway platform crowd counting, combining superior accuracy with real-time performance through physics-informed 3D tracking.

%% file: 05_Conclusion.tex
We presented a physics-constrained detect-track-count framework for real-time crowd analysis from a moving train.
By integrating a geometry-aware Phys-3D motion model with head-based detection and appearance encoding, the system achieves stable identity tracking and accurate counting under dense occlusion and camera motion.
Our experiments demonstrate that embedding physical priors into multi-object tracking substantially improves robustness and interpretability in dynamic railway environments.
We also demonstrated, that these priors are far more effective than merely increasing the complexity of the kinematic model.

The precise crowd counting capability provides multi-dimensional insights for railway station management, such as real-time safety and density management, operational efficiency and scheduling as well as capacity planning and infrastructure development.
Accurate passenger counting serves as a foundation for higher-level analytics.
Finally, future work will extend our approach to multi-camera fusion and domain adaptation for broader transportation scenarios.

\textbf{Limitations and Future Directions} While the Phys-3D framework demonstrates robust tracking and counting performance, our current dataset does not include extreme conditions such as nighttime illumination or adverse weather. Expanding data collection to encompass diverse environmental and operational scenarios would strengthen model generalization and robustness. In addition, domain randomization and simulation-based pretraining could further mitigate the data scarcity issue. Additionally, incorporating multi-modal sensing, such as LiDAR, radar, or thermal imaging, can enhance all-weather reliability and improve target discrimination in challenging visual conditions.

\section*{ACKNOWLEDGEMENTS}
This work was partly funded by the German Federal Ministry for Economic Affairs and Energy  (DeepTrain, grant no. 19S23005D).

%% file: Appendix.tex
\begin{landscape}
\section*{\uppercase{Appendix}}
\label{sec:appendix}

\subsection*{Detailed Per-Video Results}

This appendix provides the complete per-video performance results for the Phys-3D system across all 20 test videos.

\begin{table}[h]
\centering
\caption{Complete Phys-3D Detailed Performance Results Across All 20 Test Videos}
\resizebox{1.3\textwidth}{!}{%
\renewcommand{\arraystretch}{1.2}
\renewcommand{\arraystretch}{1.2}
\newcommand{\tablefont}{\scalebox{1.2}}
\begin{tabular}{|p{0.5cm}|p{0.7cm}|p{0.7cm}|p{0.7cm}|p{0.7cm}|p{0.8cm}|p{0.8cm}|p{0.8cm}|p{0.7cm}|p{0.6cm}|p{0.8cm}|p{1.0cm}|p{0.8cm}|p{0.8cm}|p{0.7cm}|p{0.8cm}|p{0.8cm}|p{0.8cm}|p{0.8cm}|p{0.8cm}|p{0.8cm}|}
\hline
{\tiny\scalebox{1.2}{\textbf{Video}}} & {\tiny\scalebox{1.2}{\textbf{MOTA}}} & {\tiny\scalebox{1.2}{\textbf{MOTP}}} & {\tiny\scalebox{1.2}{\textbf{IDF1}}} & {\tiny\scalebox{1.2}{\textbf{IDP}}} & {\tiny\scalebox{1.2}{\textbf{IDR}}} & {\tiny\scalebox{1.2}{\textbf{IDSW}}} & {\tiny\scalebox{1.2}{\textbf{Matches}}} & {\tiny\scalebox{1.2}{\textbf{FP}}} & {\tiny\scalebox{1.2}{\textbf{Misses}}} & {\tiny\scalebox{1.2}{\textbf{FAF}}} & {\tiny\scalebox{1.2}{\textbf{Precision}}} & {\tiny\scalebox{1.2}{\textbf{Recall}}} & {\tiny\scalebox{1.2}{\textbf{MT}}} & {\tiny\scalebox{1.2}{\textbf{PT}}} & {\tiny\scalebox{1.2}{\textbf{ML}}} & {\tiny\scalebox{1.2}{\textbf{LC}}} & {\tiny\scalebox{1.2}{\textbf{RC}}} & {\tiny\scalebox{1.2}{\textbf{TC}}} & {\tiny\scalebox{1.2}{\textbf{TV}}} & {\tiny\scalebox{1.2}{\textbf{MAPE}}} \\ \hline
1 & 54.26 & 19.63 & 72.96 & 81.73 & 65.89 & 2 & 86 & 16 & 41 & 0.16 & 84.62 & 68.22 & 1 & 3 & 0 & 0 & 4 & 4 & 4 & 0 \\
2 & 57.35 & 17.64 & 75.65 & 87.65 & 66.55 & 3 & 375 & 51 & 187 & 0.22 & 88.11 & 66.9 & 3 & 5 & 0 & 8 & 0 & 8 & 8 & 0 \\
3 & 57.25 & 17.66 & 73.19 & 77.53 & 69.32 & 7 & 993 & 215 & 359 & 0.52 & 82.3 & 73.58 & 12 & 10 & 3 & 24 & 0 & 25 & 25 & 0 \\
4 & 78.33 & 16.74 & 83.16 & 81.49 & 84.9 & 1 & 610 & 86 & 58 & 0.29 & 87.66 & 91.33 & 9 & 1 & 0 & 0 & 10 & 10 & 10 & 0 \\
5 & 64.42 & 16.5 & 78.33 & 79.56 & 77.14 & 3 & 874 & 175 & 208 & 0.36 & 83.37 & 80.83 & 13 & 4 & 1 & 19 & 0 & 19 & 18 & 5.56 \\
6 & 81.1 & 16.11 & 82.98 & 84.91 & 81.13 & 8 & 2440 & 196 & 319 & 0.29 & 92.59 & 88.47 & 13 & 3 & 0 & 16 & 0 & 16 & 16 & 0 \\
7 & 53.72 & 19.9 & 63.01 & 75.52 & 54.06 & 15 & 1097 & 150 & 651 & 0.29 & 88.11 & 63.07 & 6 & 10 & 5 & 0 & 21 & 21 & 21 & 0 \\
8 & 57.31 & 16.91 & 74.24 & 81.18 & 68.39 & 17 & 2522 & 473 & 1036 & 0.44 & 84.3 & 71.02 & 20 & 23 & 4 & 1 & 44 & 44 & 47 & 6.38 \\
9 & 73.84 & 15.94 & 79.76 & 88.47 & 72.6 & 87 & 10968 & 537 & 3071 & 0.26 & 95.37 & 78.26 & 71 & 51 & 2 & 0 & 123 & 123 & 124 & 0.81 \\
10 & 78.67 & 18.02 & 85.86 & 88.98 & 82.96 & 12 & 1517 & 123 & 243 & 0.19 & 92.55 & 86.29 & 14 & 5 & 0 & 19 & 0 & 19 & 19 & 0 \\
11 & 71.41 & 19.41 & 76.45 & 80.95 & 72.43 & 35 & 2507 & 266 & 596 & 0.33 & 90.53 & 81.01 & 18 & 8 & 1 & 27 & 0 & 27 & 27 & 0 \\
12 & 68.76 & 19.79 & 72.11 & 79.28 & 66.13 & 94 & 8403 & 766 & 2609 & 0.33 & 91.73 & 76.51 & 36 & 27 & 1 & 66 & 0 & 66 & 64 & 3.12 \\
13 & 76.76 & 15.26 & 83.84 & 92.13 & 76.92 & 4 & 498 & 19 & 122 & 0.08 & 96.35 & 80.45 & 8 & 4 & 0 & 11 & 0 & 12 & 12 & 0 \\
14 & 56.74 & 17.15 & 66.1 & 74.66 & 59.31 & 119 & 8217 & 1320 & 3819 & 0.7 & 86.33 & 68.58 & 34 & 58 & 7 & 0 & 100 & 100 & 100 & 0 \\
15 & 62.28 & 17.2 & 71.93 & 78.18 & 66.62 & 16 & 1947 & 296 & 688 & 0.43 & 86.9 & 74.05 & 16 & 10 & 2 & 29 & 0 & 29 & 28 & 3.57 \\
16 & 63.13 & 17.31 & 74.31 & 75.3 & 73.35 & 17 & 2850 & 602 & 694 & 0.48 & 82.65 & 80.51 & 21 & 11 & 0 & 0 & 31 & 32 & 32 & 0 \\
17 & 72.73 & 15.43 & 82.94 & 84.76 & 81.2 & 11 & 1218 & 162 & 223 & 0.34 & 88.35 & 84.64 & 11 & 4 & 1 & 0 & 15 & 15 & 16 & 6.25 \\
18 & 79.97 & 15.45 & 70.53 & 73.22 & 68.03 & 6 & 1119 & 81 & 173 & 0.12 & 93.28 & 86.67 & 3 & 3 & 0 & 6 & 0 & 6 & 6 & 0 \\
19 & 75.37 & 17.63 & 84.94 & 89.28 & 81 & 5 & 1073 & 97 & 217 & 0.34 & 91.74 & 83.24 & 15 & 7 & 2 & 0 & 21 & 21 & 24 & 12.5 \\
20 & 60.73 & 18.45 & 73.56 & 89.79 & 62.29 & 34 & 2859 & 174 & 1528 & 0.18 & 94.33 & 65.44 & 12 & 32 & 2 & 43 & 0 & 43 & 46 & 6.52 \\
Avg. & 67.21 & 17.41 & 76.29 & 82.23 & 71.51 & 24.80 & 2609 & 290 & 842 & 0.32 & 89.06 & 77.45 & 16.80 & 13.95 & 1.55 & 13.45 & 18.45 & 32.00 & 32.35 & 2.24 \\ 
\hline
\end{tabular}%
}
\label{tab:Phys5D_Complete_Detailed_Results}
\end{table}

This table provides the complete per-video analysis of Phys-3D performance across all 20 test videos. The results demonstrate consistent performance across diverse scenarios, with videos showing varying crowd densities and tracking challenges. The average values shown in the main text are calculated from these detailed per-video results. Note: LC is left counting value, RC is right counting value, and TV is the true value representing the actual number of people in the video.
\end{landscape}

% ---------------------------------------
\newpage
\onecolumn
\subsection*{Complete YOLOv11m Detector Ablation Study Results}

This section provides the complete results of the YOLOv11m detector ablation study across all 64 parameter combinations.

\begin{table}[h]
\centering
\caption{Complete YOLOv11m Detector Ablation Study Results Across All 64 Parameter Combinations}
\resizebox{0.56\textwidth}{!}{%
\renewcommand{\arraystretch}{1.2}
\newcommand{\tablefont}{\scalebox{1.2}}
\begin{tabular}{|c|c|c|c|c|c|c|c|c|}
\hline
\textbf{No.} & \textbf{Conf} & \textbf{IoU} & \textbf{Imgsz} & \textbf{mAP50} & \textbf{mAP50-95} & \textbf{Precision} & \textbf{Recall} & \textbf{Inference\_Time(ms)} \\ \hline
1 & 0.3 & 0.3 & 640 & 0.9248 & 0.6749 & 0.9272 & 0.8741 & 13.18 \\
2 & 0.3 & 0.3 & 736 & 0.9224 & 0.6871 & 0.9193 & 0.8707 & 16.86 \\
3 & 0.3 & 0.3 & 832 & 0.8997 & 0.6545 & 0.9195 & 0.8238 & 21.79 \\
4 & 0.3 & 0.3 & 960 & 0.8664 & 0.6144 & 0.8947 & 0.7786 & 29.42 \\
5 & 0.3 & 0.5 & 640 & 0.9247 & 0.6748 & 0.9246 & 0.8745 & 14.63 \\
6 & 0.3 & 0.5 & 736 & 0.9224 & 0.6869 & 0.9177 & 0.8712 & 18.18 \\
7 & 0.3 & 0.5 & 832 & 0.8997 & 0.6543 & 0.9183 & 0.8243 & 23.32 \\
8 & 0.3 & 0.5 & 960 & 0.8661 & 0.6143 & 0.9003 & 0.7736 & 31.44 \\
9 & 0.3 & 0.7 & 640 & 0.9244 & 0.6744 & 0.9316 & 0.8652 & 14.44 \\
10 & 0.3 & 0.7 & 736 & 0.9220 & 0.6865 & 0.9120 & 0.8717 & 18.05 \\
11 & 0.3 & 0.7 & 832 & 0.8991 & 0.6538 & 0.9146 & 0.8242 & 23.43 \\
12 & 0.3 & 0.7 & 960 & 0.8654 & 0.6137 & 0.8944 & 0.7741 & 31.53 \\
13 & 0.3 & 0.9 & 640 & 0.9163 & 0.6677 & 0.9263 & 0.8364 & 14.46 \\
14 & 0.3 & 0.9 & 736 & 0.9138 & 0.6801 & 0.9153 & 0.8334 & 18.22 \\
15 & 0.3 & 0.9 & 832 & 0.8892 & 0.6468 & 0.9049 & 0.7986 & 23.60 \\
16 & 0.3 & 0.9 & 960 & 0.8523 & 0.6041 & 0.8829 & 0.7413 & 31.55 \\
17 & 0.5 & 0.3 & 640 & 0.9193 & 0.6734 & 0.9411 & 0.8603 & 14.23 \\
18 & 0.5 & 0.3 & 736 & 0.9156 & 0.6851 & 0.9342 & 0.8540 & 18.05 \\
19 & 0.5 & 0.3 & 832 & 0.8946 & 0.6539 & 0.9229 & 0.8212 & 23.41 \\
20 & 0.5 & 0.3 & 960 & 0.8591 & 0.6133 & 0.9176 & 0.7569 & 31.51 \\
21 & 0.5 & 0.5 & 640 & 0.9192 & 0.6733 & 0.9390 & 0.8605 & 14.51 \\
22 & 0.5 & 0.5 & 736 & 0.9157 & 0.6850 & 0.9329 & 0.8545 & 18.18 \\
23 & 0.5 & 0.5 & 832 & 0.8946 & 0.6538 & 0.9217 & 0.8216 & 23.38 \\
24 & 0.5 & 0.5 & 960 & 0.8591 & 0.6132 & 0.9162 & 0.7571 & 31.30 \\
25 & 0.5 & 0.7 & 640 & 0.9190 & 0.6730 & 0.9355 & 0.8611 & 14.33 \\
26 & 0.5 & 0.7 & 736 & 0.9155 & 0.6847 & 0.9295 & 0.8549 & 18.25 \\
27 & 0.5 & 0.7 & 832 & 0.8942 & 0.6534 & 0.9175 & 0.8217 & 23.39 \\
28 & 0.5 & 0.7 & 960 & 0.8586 & 0.6129 & 0.9124 & 0.7574 & 31.17 \\
29 & 0.5 & 0.9 & 640 & 0.9121 & 0.6675 & 0.9263 & 0.8364 & 14.30 \\
30 & 0.5 & 0.9 & 736 & 0.9087 & 0.6796 & 0.9153 & 0.8334 & 18.11 \\
31 & 0.5 & 0.9 & 832 & 0.8859 & 0.6475 & 0.9049 & 0.7986 & 23.56 \\
32 & 0.5 & 0.9 & 960 & 0.8477 & 0.6050 & 0.8829 & 0.7413 & 31.20 \\
33 & 0.7 & 0.3 & 640 & 0.9108 & 0.6706 & 0.9572 & 0.8399 & 14.28 \\
34 & 0.7 & 0.3 & 736 & 0.9076 & 0.6832 & 0.9525 & 0.8340 & 18.18 \\
35 & 0.7 & 0.3 & 832 & 0.8845 & 0.6512 & 0.9430 & 0.7956 & 23.38 \\
36 & 0.7 & 0.3 & 960 & 0.8482 & 0.6107 & 0.9381 & 0.7276 & 31.67 \\
37 & 0.7 & 0.5 & 640 & 0.9108 & 0.6706 & 0.9563 & 0.8401 & 14.35 \\
38 & 0.7 & 0.5 & 736 & 0.9077 & 0.6831 & 0.9519 & 0.8343 & 18.09 \\
39 & 0.7 & 0.5 & 832 & 0.8846 & 0.6511 & 0.9424 & 0.7959 & 23.61 \\
40 & 0.7 & 0.5 & 960 & 0.8482 & 0.6107 & 0.9380 & 0.7276 & 31.68 \\
41 & 0.7 & 0.7 & 640 & 0.9106 & 0.6703 & 0.9543 & 0.8402 & 14.11 \\
42 & 0.7 & 0.7 & 736 & 0.9073 & 0.6829 & 0.9494 & 0.8343 & 18.11 \\
43 & 0.7 & 0.7 & 832 & 0.8845 & 0.6510 & 0.9412 & 0.7960 & 23.57 \\
44 & 0.7 & 0.7 & 960 & 0.8479 & 0.6105 & 0.9362 & 0.7278 & 31.16 \\
45 & 0.7 & 0.9 & 640 & 0.9055 & 0.6664 & 0.9216 & 0.8402 & 14.08 \\
46 & 0.7 & 0.9 & 736 & 0.9023 & 0.6791 & 0.9147 & 0.8345 & 18.11 \\
47 & 0.7 & 0.9 & 832 & 0.8782 & 0.6466 & 0.9076 & 0.7961 & 23.62 \\
48 & 0.7 & 0.9 & 960 & 0.8399 & 0.6046 & 0.8993 & 0.7284 & 30.99 \\
49 & 0.9 & 0.3 & 640 & 0.8713 & 0.6544 & 0.9844 & 0.7534 & 14.30 \\
50 & 0.9 & 0.3 & 736 & 0.8688 & 0.6671 & 0.9804 & 0.7481 & 18.09 \\
51 & 0.9 & 0.3 & 832 & 0.8426 & 0.6350 & 0.9749 & 0.7002 & 23.52 \\
52 & 0.9 & 0.3 & 960 & 0.8061 & 0.5970 & 0.9759 & 0.6277 & 31.56 \\
53 & 0.9 & 0.5 & 640 & 0.8713 & 0.6544 & 0.9842 & 0.7534 & 14.30 \\
54 & 0.9 & 0.5 & 736 & 0.8688 & 0.6671 & 0.9804 & 0.7481 & 18.09 \\
55 & 0.9 & 0.5 & 832 & 0.8426 & 0.6349 & 0.9747 & 0.7003 & 23.44 \\
56 & 0.9 & 0.5 & 960 & 0.8061 & 0.5970 & 0.9759 & 0.6277 & 31.06 \\
57 & 0.9 & 0.7 & 640 & 0.8711 & 0.6543 & 0.9836 & 0.7534 & 14.21 \\
58 & 0.9 & 0.7 & 736 & 0.8687 & 0.6671 & 0.9797 & 0.7481 & 18.11 \\
59 & 0.9 & 0.7 & 832 & 0.8426 & 0.6349 & 0.9747 & 0.7003 & 23.42 \\
60 & 0.9 & 0.7 & 960 & 0.8060 & 0.5970 & 0.9758 & 0.6277 & 31.07 \\
61 & 0.9 & 0.9 & 640 & 0.8692 & 0.6529 & 0.9748 & 0.7535 & 14.23 \\
62 & 0.9 & 0.9 & 736 & 0.8670 & 0.6659 & 0.9723 & 0.7481 & 18.14 \\
63 & 0.9 & 0.9 & 832 & 0.8403 & 0.6334 & 0.9670 & 0.7005 & 22.99 \\
64 & 0.9 & 0.9 & 960 & 0.8028 & 0.5945 & 0.9653 & 0.6278 & 29.96 \\
\hline
\end{tabular}%
}
\label{tab:YOLO11m_Complete_Ablation}
\end{table}

% ---------------------------------------
\newpage
\subsection*{Complete Virtual Counting Band Ablation Study Results}

This section provides the complete results of the virtual counting band ablation study across all parameter combinations.

\begin{table}[h]
\centering
\caption{Complete Virtual Counting Band Ablation Study Results Across All Parameter Combinations}
\resizebox{0.45\textwidth}{!}{%
\scriptsize
\renewcommand{\arraystretch}{1.2}
\begin{tabular}{|c|c|c|c|c|c|}
\hline
\textbf{Start} & \textbf{End} & \textbf{MAE} & \textbf{RMSE} & \textbf{MAPE(\%)} & \textbf{ME} \\ \hline
0.05 & 0.20 & 0.90 & 1.38 & 2.97 & -0.20 \\
0.05 & 0.15 & 1.15 & 1.75 & 4.07 & -0.95 \\
0.00 & 0.15 & 1.25 & 1.77 & 4.67 & -0.15 \\
0.10 & 0.20 & 1.40 & 2.10 & 5.28 & -1.10 \\
0.00 & 0.20 & 1.50 & 2.14 & 4.97 & 0.60 \\
0.15 & 0.30 & 1.50 & 2.37 & 8.58 & 0.20 \\
0.00 & 0.10 & 1.80 & 2.57 & 5.83 & -1.60 \\
0.10 & 0.25 & 1.80 & 2.68 & 6.94 & 0.30 \\
0.20 & 0.30 & 2.05 & 2.89 & 9.66 & -0.85 \\
0.05 & 0.25 & 2.10 & 2.93 & 6.55 & 1.10 \\
0.15 & 0.25 & 2.10 & 2.81 & 9.57 & -1.30 \\
0.20 & 0.35 & 2.40 & 3.44 & 10.61 & 1.10 \\
0.10 & 0.30 & 2.45 & 3.68 & 8.55 & 1.65 \\
0.10 & 0.15 & 2.60 & 3.65 & 9.81 & -2.50 \\
0.00 & 0.25 & 2.75 & 3.94 & 8.59 & 1.85 \\
0.05 & 0.10 & 2.75 & 3.56 & 8.78 & -2.55 \\
0.05 & 0.30 & 2.85 & 4.46 & 8.56 & 2.45 \\
0.15 & 0.35 & 2.85 & 4.42 & 11.16 & 1.95 \\
0.30 & 0.40 & 3.05 & 4.17 & 15.38 & -0.55 \\
0.15 & 0.20 & 3.15 & 4.38 & 11.45 & -3.05 \\
0.20 & 0.25 & 3.50 & 4.93 & 12.89 & -2.80 \\
0.30 & 0.35 & 3.50 & 4.43 & 16.85 & -3.40 \\
0.00 & 0.30 & 3.60 & 5.53 & 10.81 & 3.20 \\
0.10 & 0.35 & 3.90 & 6.43 & 11.85 & 3.30 \\
0.20 & 0.40 & 4.15 & 6.43 & 16.10 & 3.35 \\
0.05 & 0.35 & 4.30 & 7.44 & 11.85 & 4.10 \\
0.15 & 0.40 & 4.65 & 7.55 & 16.81 & 4.15 \\
0.00 & 0.35 & 5.00 & 8.54 & 13.79 & 4.80 \\
0.10 & 0.40 & 5.80 & 9.62 & 17.90 & 5.50 \\
0.00 & 0.05 & 5.95 & 7.77 & 18.77 & -5.95 \\
0.05 & 0.40 & 6.30 & 10.62 & 18.11 & 6.30 \\
0.00 & 0.40 & 7.00 & 11.72 & 20.05 & 7.00 \\
0.30 & 0.30 & 31.10 & 43.49 & 92.71 & -31.10 \\
0.05 & 0.05 & 31.30 & 43.69 & 93.55 & -31.30 \\
0.15 & 0.15 & 31.30 & 43.67 & 93.58 & -31.30 \\
0.10 & 0.10 & 31.35 & 43.74 & 93.66 & -31.35 \\
\hline
\end{tabular}%
}
\label{tab:Counting_Band_Complete_Ablation}
\end{table}

This comprehensive ablation study reveals the critical importance of virtual counting band configuration for accurate crowd counting. The results demonstrate that moderate-width bands (Start=0.05, End=0.20) provide optimal performance, while degenerate line settings (Start=End) cause catastrophic degradation due to sensitivity to jitter and transient detection loss. The study validates the effectiveness of the proposed virtual counting band approach over traditional line-crossing methods.

% ---------------------------------------
\newpage
\subsection*{Comparison of ImageNet Performance and Model Complexity Across CNN Architectures}

This section presents a comparative analysis of ImageNet performance and computational complexity across representative CNN backbones.

\begin{table}[h]
\centering
\caption{ImageNet performance versus computational complexity across EfficientNet and other representative CNN architectures~\cite{219}.}
\resizebox{0.99\textwidth}{!}{%
\scriptsize
\renewcommand{\arraystretch}{1.2}
\begin{tabular}{|l|c|c|c|c|c|c|}
\hline
\textbf{Model} & \textbf{Top-1 Acc.} & \textbf{Top-5 Acc.} & \textbf{\#Params} & \textbf{Ratio-to-EfficientNet} & \textbf{\#FLOPs} & \textbf{Ratio-to-EfficientNet} \\ \hline
EfficientNet-B0 & 77.1\% & 93.3\% & 5.3M & 1.0× & 0.39B & 1.0× \\ \hline
ResNet-50 & 76.0\% & 93.0\% & 26M & 4.9× & 4.1B & 11× \\ \hline
DenseNet-169 & 76.2\% & 93.2\% & 14M & 2.6× & 3.5B & 8.9× \\ \hline
EfficientNet-B1 & 79.1\% & 94.4\% & 7.8M & 1.0× & 0.70B & 1.0× \\ \hline
ResNet-152 & 77.8\% & 93.8\% & 60M & 7.6× & 11B & 16× \\ \hline
DenseNet-264 & 77.9\% & 93.9\% & 34M & 4.3× & 6.0B & 8.6× \\ \hline
Inception-v3 & 78.8\% & 94.4\% & 24M & 3.0× & 5.7B & 8.1× \\ \hline
Xception & 79.0\% & 94.5\% & 23M & 3.0× & 8.4B & 12× \\ \hline
EfficientNet-B2 & 80.1\% & 94.9\% & 9.2M & 1.0× & 1.0B & 1.0× \\ \hline
Inception-v4 & 80.0\% & 95.0\% & 48M & 5.2× & 13B & 13× \\ \hline
Inception-ResNet-v2 & 80.1\% & 95.1\% & 56M & 6.1× & 13B & 13× \\ \hline
EfficientNet-B3 & 81.6\% & 95.7\% & 12M & 1.0× & 1.8B & 1.0× \\ \hline
ResNeXt-101 & 80.9\% & 95.6\% & 84M & 7.0× & 32B & 18× \\ \hline
PolyNet & 81.3\% & 95.8\% & 92M & 7.7× & 35B & 19× \\ \hline
EfficientNet-B4 & 82.9\% & 96.4\% & 19M & 1.0× & 4.2B & 1.0× \\ \hline
SENet & 82.7\% & 96.2\% & 146M & 7.7× & 42B & 10× \\ \hline
NASNet-A & 82.7\% & 96.2\% & 89M & 4.7× & 24B & 5.7× \\ \hline
AmoebaNet-A & 82.8\% & 96.1\% & 87M & 4.6× & 23B & 5.5× \\ \hline
PNASNet & 82.9\% & 96.2\% & 86M & 4.5× & 23B & 6.0× \\ \hline
EfficientNet-B5 & 83.6\% & 96.7\% & 30M & 1.0× & 9.9B & 1.0× \\ \hline
AmoebaNet-C & 83.5\% & 96.5\% & 155M & 5.2× & 41B & 4.1× \\ \hline
EfficientNet-B6 & 84.0\% & 96.8\% & 43M & 1.0× & 19B & 1.0× \\ \hline
EfficientNet-B7 & 84.3\% & 97.0\% & 66M & 1.0× & 37B & 1.0× \\ \hline
GPipe & 84.3\% & 97.0\% & 557M & 8.4× & -- & -- \\ \hline
\end{tabular}%
}
\label{tab:efficientnet_comparison}
\end{table}

As shown in Table~\ref{tab:efficientnet_comparison}, EfficientNet\mbox{-}B0 achieves a favorable balance between accuracy and computational complexity. With only 5.3M parameters and 0.39~GFLOPs, it delivers competitive ImageNet performance while substantially reducing resource requirements. This efficiency–accuracy trade-off makes it an ideal backbone for real-time ReID tasks in multi-camera and edge computing environments.

% ---------------------------------------
\newpage
\subsection*{MOT Evaluation Dataset: 20 Videos}

For comprehensive evaluation of the multi-object tracking and counting system, we select a representative subset of 20 video sequences from the complete MOT-RPCH dataset. This benchmark covers diverse operational conditions, including multiple resolutions (1536×864 to 2304×1296), frame rates (25–59.94 FPS), and target scales (AvgH$\approx$49 px). The evaluation set comprises 18,548 frames, 647 identities, and 73,799 bounding boxes, providing a robust basis for performance assessment using standard MOT and counting metrics.

\begin{table}[h]
\centering
\caption{Twenty video sequences selected from MOT-RPCH for multi-object tracking and counting evaluation.}
\resizebox{0.99\textwidth}{!}{%
\scriptsize
\renewcommand{\arraystretch}{1.3}
\begin{tabular}{|c|c|c|c|c|c|c|c|c|c|c|c|c|c|c|}
\hline
\textbf{Video} & \textbf{Resolution} & \textbf{FPS} & \textbf{Duration (s)} & \textbf{Frames} & \textbf{Frame Range} & \textbf{BBox Count} & \textbf{Objects} & \textbf{MinH (px)} & \textbf{MaxH (px)} & \textbf{MinW (px)} & \textbf{MaxW (px)} & \textbf{AvgH (px)} & \textbf{AvgW (px)} & \textbf{Format} \\ \hline
1 & 1920×1080 & 25 & 8.2 & 205 & 98–196 & 129 & 4 & 27.24 & 57.35 & 20.38 & 52.67 & 37.75 & 30.39 & .mp4 \\ \hline
2 & 1920×1080 & 25 & 11.44 & 286 & 41–267 & 565 & 8 & 20.19 & 86.36 & 16.63 & 69.53 & 40.08 & 31.74 & .mp4 \\ \hline
3 & 1920×1080 & 25 & 18.12 & 453 & 36–451 & 1359 & 25 & 14.64 & 153.88 & 16.36 & 106.39 & 52.36 & 41.48 & .mp4 \\ \hline
4 & 2304×1296 & 59.94 & 7.84 & 470 & 44–427 & 670 & 10 & 34.95 & 130.56 & 24.32 & 129.21 & 63.99 & 55.43 & .mp4 \\ \hline
5 & 2304×1296 & 59.94 & 11.29 & 677 & 73–639 & 1085 & 18 & 24.82 & 197.75 & 17.58 & 227.34 & 70.60 & 65.51 & .mp4 \\ \hline
6 & 2304×1296 & 59.94 & 11.18 & 670 & 2–666 & 2911 & 16 & 18.00 & 289.02 & 20.16 & 259.88 & 77.29 & 64.89 & .mp4 \\ \hline
7 & 2304×1296 & 59.94 & 10.59 & 635 & 38–615 & 1807 & 21 & 26.90 & 146.26 & 18.23 & 153.21 & 57.48 & 52.73 & .mp4 \\ \hline
8 & 1536×864 & 59.94 & 24.96 & 1496 & 146–1215 & 3575 & 47 & 11.61 & 121.28 & 9.15 & 123.57 & 38.45 & 35.60 & .mp4 \\ \hline
9 & 1536×864 & 59.94 & 35.64 & 2136 & 12–2038 & 17243 & 124 & 10.90 & 118.12 & 7.05 & 129.81 & 33.12 & 30.31 & .mp4 \\ \hline
10 & 1536×864 & 59.94 & 13.48 & 808 & 5–790 & 1804 & 19 & 16.99 & 147.95 & 11.01 & 149.34 & 36.14 & 32.12 & .mp4 \\ \hline
11 & 1536×864 & 59.94 & 13.85 & 830 & 1–811 & 3317 & 27 & 19.80 & 119.80 & 13.53 & 117.22 & 38.06 & 33.33 & .mp4 \\ \hline
12 & 1536×864 & 59.94 & 39.44 & 2364 & 13–2346 & 11474 & 64 & 16.06 & 125.98 & 12.54 & 160.40 & 37.65 & 34.86 & .mp4 \\ \hline
13 & 2304×1296 & 29.97 & 11.48 & 344 & 21–344 & 624 & 12 & 26.93 & 142.54 & 18.41 & 146.67 & 55.66 & 48.32 & .mp4 \\ \hline
14 & 2304×1296 & 59.94 & 33.38 & 2001 & 92–1988 & 12249 & 100 & 16.18 & 183.24 & 13.09 & 205.83 & 53.81 & 51.92 & .mp4 \\ \hline
15 & 1536×864 & 59.94 & 19.19 & 1150 & 343–1077 & 2651 & 28 & 12.78 & 104.00 & 9.62 & 119.77 & 37.33 & 34.88 & .mp4 \\ \hline
16 & 1536×864 & 59.94 & 24.07 & 1443 & 178–1443 & 3561 & 32 & 15.15 & 115.60 & 12.86 & 118.58 & 41.97 & 38.90 & .mp4 \\ \hline
17 & 1536×864 & 59.94 & 8.51 & 510 & 14–510 & 1494 & 16 & 20.40 & 120.98 & 14.72 & 120.92 & 43.05 & 38.02 & .mp4 \\ \hline
18 & 2304×1296 & 59.94 & 12.01 & 720 & 1–720 & 1303 & 6 & 30.38 & 204.36 & 21.60 & 222.35 & 66.81 & 57.20 & .mp4 \\ \hline
19 & 2304×1296 & 59.94 & 5.09 & 305 & 16–299 & 1341 & 24 & 24.39 & 167.54 & 20.14 & 154.06 & 56.02 & 52.94 & .mp4 \\ \hline
20 & 2304×1296 & 59.94 & 17.43 & 1045 & 53–1045 & 4637 & 46 & 14.33 & 116.91 & 10.62 & 131.26 & 44.73 & 43.78 & .mp4 \\ \hline
\textbf{Total} & -- & -- & \textbf{337.19} & \textbf{18548} & -- & \textbf{73799} & \textbf{647} & \textbf{20.13} & \textbf{142.47} & \textbf{15.40} & \textbf{144.90} & \textbf{49.12} & \textbf{43.72} & -- \\ \hline
\end{tabular}%
}
\label{tab:20MOTValVideos}
\end{table}
\vspace{2mm}

% ---------------------------------------
% \newpage
\subsection*{ReID Training and Ablation Dataset: 7 Videos}

We trained the ReID model using seven video sequences. This subset provides diverse appearance variations and sufficient identity samples for robust feature learning, comprising 6,240 frames and 238 tracked identities.

\begin{table}[h]
\centering
\caption{Seven video sequences for ReID training and ablation studies.}
\resizebox{0.99\textwidth}{!}{%
\scriptsize
\renewcommand{\arraystretch}{1.3}
\begin{tabular}{|c|c|c|c|c|c|c|c|c|c|c|c|c|c|c|}
\hline
\textbf{Video} & \textbf{Resolution} & \textbf{FPS} & \textbf{Duration (s)} & \textbf{Frames} & \textbf{Frame Range} & \textbf{BBox Count} & \textbf{Objects} & \textbf{MinH (px)} & \textbf{MaxH (px)} & \textbf{MinW (px)} & \textbf{MaxW (px)} & \textbf{AvgH (px)} & \textbf{AvgW (px)} & \textbf{Format} \\ \hline
1 & 1920×1080 & 25 & 7.24 & 181 & 67–166 & 437 & 8 & 22.44 & 103.24 & 17.95 & 87.47 & 46.75 & 36.23 & .mp4 \\ \hline
2 & 1920×1080 & 25 & 7.16 & 179 & 64–151 & 181 & 3 & 22.85 & 58.42 & 14.31 & 49.01 & 36.74 & 27.28 & .mp4 \\ \hline
3 & 2304×1296 & 59.94 & 11.54 & 692 & 51–664 & 1914 & 53 & 15.91 & 194.65 & 14.42 & 221.60 & 57.45 & 51.93 & .mp4 \\ \hline
4 & 2560×1440 & 59.94 & 6.99 & 419 & 90–419 & 905 & 18 & 12.72 & 127.42 & 10.28 & 117.93 & 26.46 & 24.48 & .mp4 \\ \hline
5 & 2560×1440 & 59.94 & 45.71 & 2740 & 1–2697 & 7415 & 93 & 10.84 & 122.27 & 8.69 & 127.90 & 29.51 & 29.76 & .mp4 \\ \hline
6 & 2304×1296 & 59.94 & 15.01 & 900 & 22–898 & 1171 & 9 & 26.70 & 214.66 & 21.27 & 179.46 & 68.55 & 55.99 & .mp4 \\ \hline
7 & 2304×1296 & 59.94 & 18.84 & 1129 & 1–724 & 3265 & 54 & 16.65 & 162.63 & 10.00 & 199.17 & 56.05 & 54.55 & .mp4 \\ \hline
\textbf{Total} & -- & -- & \textbf{112.49} & \textbf{6240} & -- & \textbf{15288} & \textbf{238} & \textbf{18.30} & \textbf{140.47} & \textbf{13.85} & \textbf{140.36} & \textbf{45.93} & \textbf{40.03} & -- \\ \hline
\end{tabular}%
}
\label{tab:ReIDDataDistribution}
\end{table}